\documentclass{article}

    \usepackage[preprint]{neurips_2022}

\usepackage[utf8]{inputenc} 
\usepackage[T1]{fontenc}    
\usepackage{hyperref}       
\usepackage{url}            
\usepackage{booktabs}       
\usepackage{amsfonts}       
\usepackage{nicefrac}       
\usepackage{microtype}      
\usepackage[dvipsnames]{xcolor}
\usepackage{comment}
\usepackage{graphicx}
\usepackage{wrapfig, blindtext, booktabs}
\usepackage{caption}
\usepackage{subcaption}
\usepackage[most]{tcolorbox}
\tcbuselibrary{listings}
\newtcblisting{codeblock}[1]{title=#1}
\usepackage{tabulary}
\usepackage{wrapfig}
\usepackage{longtable}
\usepackage{amsmath}
\usepackage{blindtext}
\newcommand\blfootnote[1]{%
\begingroup
\renewcommand\thefootnote{}\footnote{#1}%
\addtocounter{footnote}{-1}%
\endgroup
}

\newcount\Comments  
\definecolor{purple}{rgb}{0.5,0,1}
\definecolor{teal}{rgb}{0.33,0.65,0.55}
\definecolor{green}{rgb}{0.1,0.65,0.1}
\newcommand{\kibitz}[2]{\ifnum\Comments=1\textcolor{#1}{#2}\fi}

\title{Synthetic Imitation Edit Feedback for \\ Factual Alignment in Clinical Summarization}

\author{%
  Prakamya Mishra\thanks{indicates equal contribution} $^{1}$, 
  Zonghai Yao\footnotemark[1] $^{1}$\\ 
  \bf{Shuwei Chen}$^{2}$, 
  \bf{Beining Wang}$^{2}$, 
  \bf{Rohan Mittal} $^{1}$, 
  \bf{Hong Yu}$^{1,3}$
  \\
  University of Massachusetts, Amherst$^1$, Fudan University$^2$\\
  University of Massachusetts, Lowell$^3$
  \\
  \texttt\ \{\href{mailto:prakamyamish@umass.edu}{\textcolor{BlueViolet}{prakamyamish}}, \href{mailto:zonghaiyao@umass.edu}{\textcolor{BlueViolet}{zonghaiyao}}\}@\textcolor{BlueViolet}{umass.edu}
  \\
}

\begin{document}

\maketitle

\begin{abstract}

Large Language Models (LLMs) like the GPT and LLaMA families have demonstrated exceptional capabilities in capturing and condensing critical contextual information and achieving state-of-the-art performance in the summarization task. 
However, community concerns about these models' hallucination issues continue to rise. LLMs sometimes generate factually hallucinated summaries, which can be extremely harmful in the clinical domain NLP tasks (e.g., clinical note summarization), where factually incorrect statements can lead to critically erroneous diagnoses. 
Fine-tuning LLMs using human feedback has shown the promise of aligning LLMs to be factually consistent during generation, but such training procedure requires high-quality human-annotated data, which can be extremely expensive to get in the clinical domain. 
In this work, we propose a new pipeline using ChatGPT\footnote{We used ChatGPT using Azure OpenAI Service: https://azure.microsoft.com/en-us/products/ai-services/openai-service/} instead of human experts to generate high-quality feedback data for improving factual consistency in the clinical note summarization task.
We focus specifically on edit feedback because recent work discusses the shortcomings of human alignment via preference feedback in complex situations (such as clinical NLP tasks that require extensive expert knowledge), as well as some advantages of collecting edit feedback from domain experts. 
In addition, although GPT has reached the expert level in many clinical NLP tasks (e.g., USMLE QA), there is not much previous work discussing whether GPT can generate expert-level edit feedback for LMs in the clinical note summarization task. We hope to fill this gap. 
Finally, our evaluations demonstrate the potential use of GPT edits in human alignment, especially from a factuality perspective.\footnote{Our code and dataset will be released at \url{https://github.com/seasonyao/LearnFromHumanEdit}}

\blfootnote{To appear in NeurIPS 2023 Workshop SyntheticData4ML}
 
\end{abstract}

\section{Introduction}


\begin{figure}[ht]
  \centering
    \includegraphics[width=1\textwidth]{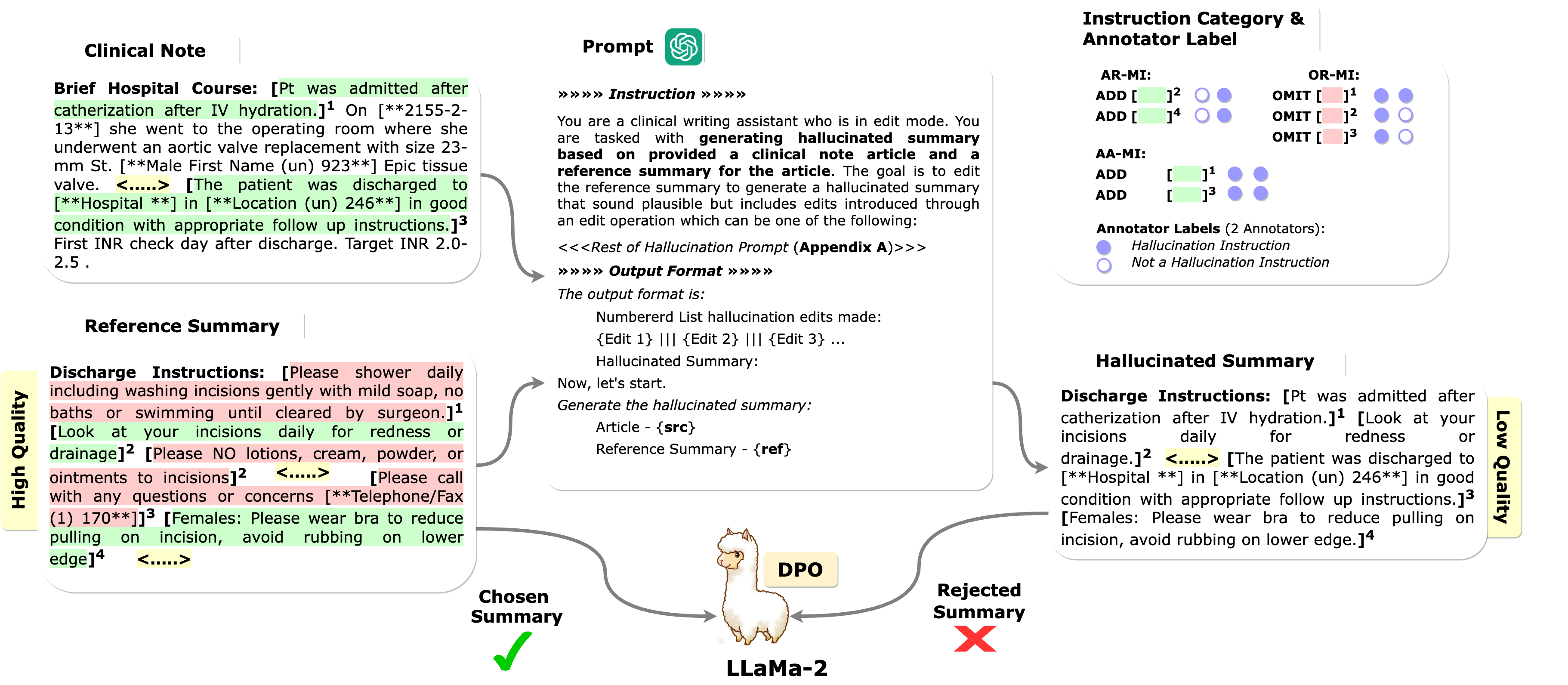}
  \caption{Illustration of the proposed synthetic imitation edit data generation and training pipeline. Given a clinical note and the corresponding reference summary (factually high-quality summary), we prompt ChatGPT to generate a hallucinated summary (factually low-quality summary) by generating edit instructions to the \textbf{ADD} hallucinated sentences and \textbf{OMIT}  critically important sentences from the reference summary \& the clinical note. The high \& low-quality summary pairs are used to train LLaMA-2 using DPO, where the reference high-quality summary is the chosen one and the hallucinated low-quality summary is the rejected one. In the figure, the sentences with \colorbox{green!20}{green} background are the edits to be included in the hallucinated summary, whereas the sentences with \colorbox{red!20}{red} background are not. The annotated instruction category and hallucination labels corresponding to the ChatGPT generated edit instructions are also shown.}
  \label{fig:main}
\end{figure}

Large language models (LLMs) such as GPT-3 \cite{NEURIPS2020_1457c0d6}, GPT-4 \cite{openai2023gpt4}, PaLM \cite{chowdhery2022palm}, LLaMA \cite{touvron2023llama}, and LLaMA-2 \cite{touvron2023llama} have significantly advanced generative AI, outperforming smaller models like T5 \cite{10.5555/3455716.3455856} and GPT-2 \citep{radford2019language} in both language understanding and natural language generation capabilities, as demonstrated in recent studies \cite{brown2020language,Sanh2021,chowdhery2022palm,longpre2023flan,openai2023gpt4,Yang2023.10.26.23297629}.
However, these LLMs still have an inherent tendency to generate hallucinations producing factually inconsistent outputs, hurting their reliability in being helpful, truthful, and harmless \citep{10.1145/3571730, openai2023gpt4, zhang2023language, maynez-etal-2020-faithfulness}. Recently, alignment of these LLMs to generate factually consistent summaries has been an active area of research, especially in the generation tasks \citep{pmlr-v202-shi23a, cao-etal-2022-hallucinated, kang-hashimoto-2020-improved, goyal-etal-2023-factual}, where human alignment with different optimization techniques like RLHF \citep{NEURIPS2022_b1efde53, ziegler2020finetuning, NEURIPS2020_1f89885d}, RLAIF \citep{lee2023rlaif}, RRHF \citep{yuan2023rrhf}, RAFT \citep{teed2020raft} using PPO \citep{baselines}, DPO \citep{rafailov2023direct} and SALT \citep{yao2023improving} have demonstrated their effectiveness in aligning these hallucinated models. Despite the effectiveness of these alignment methods, they require a significant amount of human-annotated data to illustrate human preference during training, which can be scarce in clinical domains where such expert-level annotations can be expensive to get. Synthetically generated data is used in such domains where there is a lack of preference-based human-annotated data \citep{li2023feasibility, yoo-etal-2021-gpt3mix-leveraging, dai2023auggpt}.

In this work, we focus on the clinical note summarization task in the clinical domain. Clinical notes consist of a patient's history throughout their visit to a hospital, which includes nursing \& physician notes, ECG reports, radiology reports, and etc. The clinical note summarization task is the task to generate a discharge summary from a clinical note that can later be used for the correct diagnosis of the patient based on the visit history present in the clinical note. 
Using LLMs to generate such discharge summaries could drastically improve the efficiency in the diagnosis of a patient's history. However, due to the hallucinations generated by LLMs, their generated outputs are not reliable, resulting in the need for alignment training using preference-based data. 

Human feedback for summarization can come in different forms. Comparison-based feedback, Scalar feedback, and Label feedback are more common and well-studied types for NLP tasks \citep{stiennon2020learning, ziegler2019fine, zhu2023principled, biyik2019asking, wilde2021learning, freedman2021choice}. Edit feedback and Language feedback are two more informative but harder-processed types \citep{casper2023open}. Recent works \citep{casper2023open, ji2023ai, yao2023improving} discussed some limitations of current common feedback types and the advantages of adding Edit or Language feedback for better human alignment, especially for tasks that need expert domain knowledge.
However, it is challenging to collect real-world doctor edit feedback due to privacy protection and strict data regulations like HIPAA ~\cite{rindfleisch1997privacy, annas2003hipaa}.
Generating a synthetic imitation edit feedback dataset is one potential solution. Manually constructing a large amount of such synthetic data using domain experts is time and effort-consuming~\cite {kelly2019key, abacha2023empirical}.
An alternative is to build the synthetic dataset by leveraging large language models such as ChatGPT~\cite{eysenbach2023role, li2023feasibility, Dai2023ChatAugLC}.

In this work, we mainly focus on previously less-studied edit feedback data to better align LMs to generate factually correct clinical note summaries. 
Although GPT has reached the expert level in many clinical NLP tasks (e.g., USMLE QA) \citep{kung2023performance, gilson2023does}, there is not much previous work discussing whether GPT can generate expert-level edit feedback for LMs in the clinical NLP tasks. Specifically, we propose to use high-quality synthetically LLMs-generated edit feedback for fine-tuning LLMs and LMs using the recent SOTA alignment methods DPO \citep{rafailov2023direct} for improving factuality in the model-generated summaries.

In this work, we demonstrate how high-quality synthetic edit-feedback can be used to improve the factual alignment of the LLMs in the clinical domain for the clinical note summarization task. We propose a new pipeline to generate synthetic preference-based data. We use ChatGPT to add factual hallucinations to generate factuality-based low-quality summaries given the original factuality-based high-quality reference summaries and the corresponding clinical notes. We then treat the high-quality summaries as the DPO-chosen ones and the low-quality summaries as the DPO-rejected ones used for the synthetic edit feedback. Our experiments demonstrate the effectiveness of the synthetic edit-feedback for improving factuality in the model-generated summaries of the LLaMA-2 7B (4.26\%$\uparrow$ Rouge-1, 4.67\%$\uparrow$ Factuality) and GPT2 (5.97\%$\uparrow$ Rouge-1, 8\%$\uparrow$ Factuality) model when trained using DPO compared to simple fine-tuning (SFT). Outputs from models trained using our approach also received a higher preference for factuality by human annotators.

\section{Related Work}
Recent research has yielded promising findings regarding the application of Large Language Models (LLMs) for data augmentation, particularly in tasks like summarization, translation, and code generation \cite{li2023feasibility, Dai2023ChatAugLC, zhou2022large, dai2022promptagator, yoo2021gpt3mix}. 
\cite{gilardi2023chatgpt} and\cite{ding2022gpt} have investigated the efficacy and precision of LLMs in the context of data annotation. Their findings have demonstrated promising results, showcasing the potential of LLMs to perform on par with or even surpass human Crowd-Workers in terms of accuracy.
\cite{bonifacio2022inpars} employed LLMs to generate positive sample pairs for the training of downstream models. 
Additionally, within the biomedical domain, recent research endeavors have explored the capabilities of LLMs in clinical text mining and doctor-patient conversation tasks, aiming to tackle issues related to suboptimal performance and privacy concerns \citep{tang2023does, abacha2023overview, wang2023umass_bionlp, wang2023notechat, liao2023differentiate}.

On the other hand, \cite{10.5555/3495724.3495977} points out that standard Supervised Fine-tuning (SFT) makes important errors (such as hallucinations) and unimportant errors (such as minor grammatical errors) have the same impact on the final loss, which leaves the model still unable to consistently produce output of human-determined high quality (such as factuality).
Recent work has shown the promise of learning with human feedback paradigms to produce human-determined high-quality text \cite{bohm2019better, ziegler2019fine, stiennon2020learning, akyurek2023rl4f, dong2023raft, zhao2023slic, yuan2023rrhf}.
In the clinical domain, a lack of medical knowledge often results in LMs and LLMs generating a large number of factual errors~\cite{petroni2019language, sung2021can, yao2022extracting, yao2022context}. In this paper, we will focus on using feedback learning to improve factuality.

Recent works discussed some limitations of current common feedback types and the advantages of adding edit or language feedback for better human alignment, especially for tasks that need expert domain knowledge \citep{casper2023open, yao2023improving}.
\citep{casper2023open} discussed the pros and cons of five different feedback types: comparison-based feedback, scalar feedback, label feedback, edit feedback, and language feedback, and suggested some more exploration for less-studied types like edit feedback and language feedback.
\cite{yao2023improving} showed edit feedback is a natural way to collect feedback from doctors in workflows where they may be working off of an AI-generated summary in their workflow.
This makes exploring edit feedback in the clinical domain attractive: when experts don’t have time to spend extra time providing a large amount of feedback data to the research community, how can we use the data from their daily work to help human alignment?
However, due to privacy reasons, the content of the doctor's daily work cannot be released for research usage.
In the work, we hope to explore how to use ChatGPT, which has reached an expert level in the medical license exam \citep{kung2023performance, gilson2023does}, to generate a large number of synthetic imitation edits. And then, we used it to train smaller LLMs such as Llama2 \citep{touvron2023llama}.


\section{Problem Statement}

Given an available dataset $D\mathbin{:}\{X,Y_{+}\}$ of $C$ clinical notes $X\mathbin{:}\{x^1,x^2,...x^C\}$, their corresponding ground truth reference discharge summaries $Y_{+}:\{y_{+}^1,y_{+}^2,...y_{+}^C\}$, and a reference model $\pi_{ref}$, the aim of the clinical note summarization task $T$ is to train the model $\pi_{ref}(y_{+}^i|x^i)$. Here the $i^{th}$ clinical note $x^i\mathbin{:}\{x_1^{i},x_2^{i},...x_n^{i}\}$ consists of $n$ tokens ($j^{th}$ token represented by $x_j^{i}$) and the $i^{th}$ reference summary $y_{+}^i\mathbin{:}\{y_{+,1}^{i},y_{+,2}^{i},...y_{+,m}^{i}\}$ consists of $m$ tokens ($j^{th}$ token represented by $y_{+,j}^{i}$ \& $m<<n$). The standard way to fine-tune $\pi^{ref}$ on $T$ is to simply fine-tune $\pi^{ref}$ using the cross-entropy loss $\ell_ce(y_{+}^i,\pi^{ref}(x^i))$  over the original training dataset $D$. The model fine-tuned using this approach is represented by $\pi_{sft}$.

Aligning $\pi_{ref}$ using DPO requires the need for preference-based data $D_{pref}\mathbin{:}\{X,Y_{+},Y_{-}\}$, where $Y_{+}$ is a set of preferred summaries, and $Y_{-}$ are the dispreferred ones ($Y_{-}:\{y_{-}^1,y_{-}^2,...y_{-}^k\}$). Such a preference is usually gathered through human annotation or is generated synthetically. As previously explained, not only gathering human annotations is expensive in the clinical domain, but even generating synthetic data using standard approaches like corruption \citep{chen2023purr} can be challenging.

So in this work, we (1) Propose a new pipeline to generate high-quality synthetic edit data $Y_{-}$ from $Y_{+}$ to imitate  $D_{pref}$, where $Y_{-}$ acts as the dispreferred summary; (2) Use the synthetically generated preference data $D_{pref}\mathbin{:}\{X,Y_{+},Y_{-}\}$ to fine-tune $\pi_{ref}$ using DPO to align $\pi_{\theta}$ to generate factually consistent outputs, where $\pi_{\theta}$ is the model that is being trained. In the following subsections, we describe the synthetic edit feedback generation pipeline and the training procedure in detail.

\subsection{Synthetic Imitation Edit Feedback} \label{feedback_description}
For summarization alignment using DPO, the model learns from the preference data $D_{pref}$ by learning to increase the likelihood of the preferred summaries $Y_{+}$ and to decrease the likelihood of the dispreferred summaries $Y_{-}$. Usually, $Y_{+}$ is easy to get from the ground truth labels (reference summaries) in $D$, but on the other hand, $Y_{-}$ is usually hard to get. Synthetic data generating using any corruption function $f_c(Y_{+})$ to generate $Y_{-}$ from $Y_{+}$ have been explored to generate the corrupted summaries, which can be used to act as dispreferred summaries. However leveraging a corruption-based synthetic preference data generation approach for factuality alignment can be challenging and misaligned to the final factuality alignment objective, especially in the clinical domain where some phrases (clinical instructions) in the discharge summaries are extremely important for the correct diagnosis of the patient. So using incorrectly (corrupting incorrect phrases in the summaries of $Y_{+}$) corrupted $Y_{+}$ to act as $Y_{-}$ for aligning the model using DPO can lead to poor factuality alignment.

In order to generate correct high-quality synthetic edit data for factuality alignment, we propose to use off-the-shelf LLMs like ChatGPT to act as a corruption function $f_h$ that can be used to generate imitation edit data $Y_{-}$ by adding hallucination to $Y_{+}$. In order to do so, we prompt ChatGPT to generate a hallucinated summary $f_h\mathbin{:}\{x^i,y_{+}^i\}\rightarrow y_{-}^i$ given a clinical note $x^i$ and the corresponding reference summary $y_{+}^i$, as shown in Figure \ref{fig:main}. The detailed prompt is attached in \textbf{Appendix \ref{appendixA}} Table \ref{fig:edit-prompt}. For $f_h$, we use the clinical note $x^i$ and the corresponding reference summary $y_{+}^i$ to generate a hallucinated summary $y_{-}^i$ that sounds plausible but includes edits introduced through the edit operations listed below:
\begin{itemize}
    \item \textbf{ADD Operation}: Intentionally add medico-legally essential words from the article required for accurate diagnosis and treatment documentation.
    \item \textbf{OMIT Operation}: Intentionally omit medico-legally essential words in the reference summary required for accurate diagnosis and treatment documentation.
\end{itemize}
For generating the hallucinated summaries $y_{-}^i$'s in $Y_{-}$, we prompt ChatGPT ($f_h$) to first list a set of $I$ edit instructions $E^i\mathbin{:}\{e_{1}^i,e_{2}^i,...e_{I}^i\}$, where each instruction consists of either an ADD or OMIT operation on contents in the clinical note $X^i$ or the corresponding reference summary $Y_{+}^j$, Then the prompt uses the generated instructions in $E^i$ to generate the hallucinated summary $y_{-}^i$. Since we prompt ChatGPT to add/omit medico-legally unimportant/important information respectively, resulting in a decrease in the factual consistency of the content in the hallucinate summary $y_{-}^i$, so we treat $y_{-}^i$ as the dispreferred summary and the $y_{+}^i$ as the preferred one for generating the preference data $D_{pref}$. Since edit operations (ADD/OMIT) used in $f_h$ generate $Y_{-}$ that are of low quality in terms of clinical factuality, so in line with the final objective of clinical factuality alignment, we treat $Y_{-}$ as a set of dispreferred summaries and the reference summaries $Y_{+}$ as a set of preferred ones (since they are the ground truth and are of higher quality with respect to factuality). Examples of the generated hallucination edit instructions $E^i$ along with the hallucinated summaries $Y_{-}$ are attached in the \textbf{Appendix \ref{appendixD}}.

\subsection{Preference-based Training}

\begin{equation}
\label{equ:dpo}
\ell_{dpo}(\pi_{\theta};\pi_{ref})=-\mathbin{E}_{(x^i,y_{+}^i,y_{-}^i)\sim D_{pref}}\left(\log \sigma\left[\beta\log\frac{\pi_{\theta}(y_{+}^i|x^i)}{\pi_{ref}(y_{+}^i|x^i)}-\beta\log\frac{\pi_{\theta}(y_{-}^i|x^i)}{\pi_{ref}(y_{-}^i|x^i)}\right]\right)
\end{equation}

For aligning $\pi_{ref}$ using DPO ($\pi_{ref}\rightarrow\pi_{\theta}$), we train the model by optimizing the loss function $\ell_{dpo}$ shown in equation \ref{equ:dpo}, where given the preference data $D_{pref}\mathbin{:}\{X,Y_{+},Y_{-}\}$ consisting of a set of clinical notes $x^i$, preferred summaries $Y_{+}^i$ (ground truth reference summaries), and the dispreferred summaries $Y_{-}^i$ (hallucinated summaries), the model learns to increase the likelihood of the preferred summaries $Y_{+}^i$ and to decrease the likelihood of the dispreferred summaries $Y_{-}^i$. In the equation, $\pi_{ref}$ is the base model and $\pi_{\theta}$ is the model being trained to have improved alignment and $\beta$ is used to scale the weight on how incorrect the model should treat the dispreferred summary $y_{-}^i$ relative to the preferred summary $y_{+}^i$. The higher the $\beta$ beta, the less the divergence from the initial policy $\pi_{ref}$.

\begin{figure}[ht]
  \centering
    \includegraphics[width=0.7\textwidth]{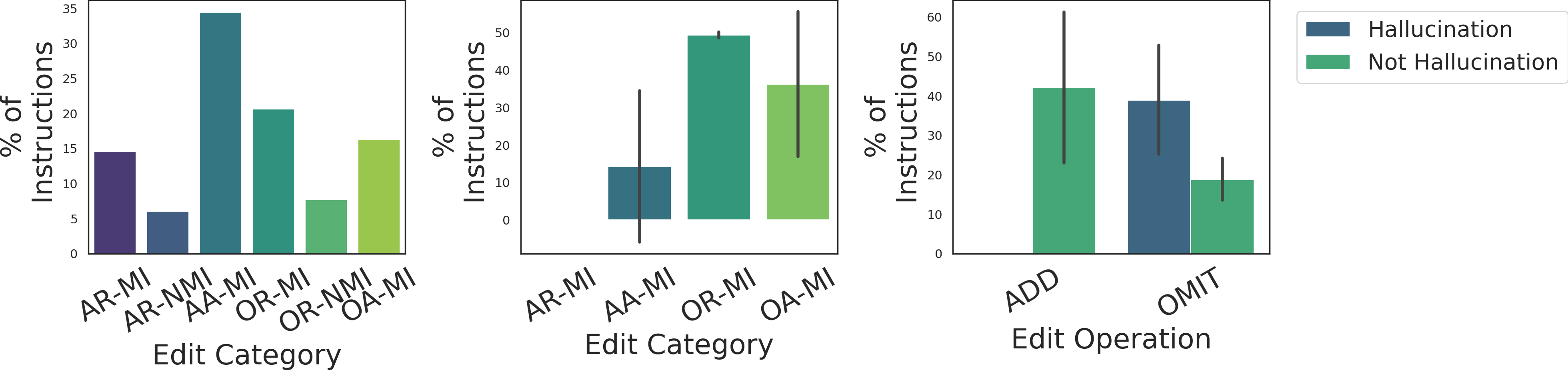}
  \caption{\textbf{First:} \% of edits made in the human evaluation samples for each edit type listed in Table \ref{tab:edit_types}. \textbf{Second:} \% of edits made by the ADD/OMIT edits mentioned in the generated instructions that resulted in hallucinations according to the annotators. \textbf{Third:} \% of the ADD/OMIT edits for generating hallucinated/non-hallucinated edits summaries according to the annotators. Annotation for all the instructions from our annotators had a mean Kappa score $k$ of 0.38 (\textbf{Appendix \ref{appendixD}}).}
  \label{fig:HE}
  \vspace{-5mm}
\end{figure}

\section{Results}
In this section, we first evaluate the quality of the synthetic edit data generated by our pipeline, where we evaluate the correctness of hallucination edits and the diversity of edits made by our pipeline. 
Then we conduct experiments for external evaluation on the downstream summarization tasks. Following previous work \citep{cai-etal-2022-generation}, we used the discharge instructions from the MIMIC-III database \citep{johnson2016mimic} in our experiments for clinical note summarization, consisting of 25k/3k/3k train/valid/test respective clinical note and reference summaries. MIMIC-III is a publicly available repository of de-identified health records of over 40,000 patients collected from the Beth Israel Deaconess Medical Center in Massachusetts. Due to resource limitations, we restricted the train/valid/test set to 5k/128/128.


\subsection{Synthetic Edit Feedback Evaluation} \label{section4_1}


\setlength\intextsep{1pt}
\begin{wraptable}{r}{7.5cm}
\centering
\small
\caption{Hallucination Edit Types}
\resizebox{0.5\textwidth}{!}{%
\begin{tabular}{cll}
\hline
\multicolumn{1}{c|}{\begin{tabular}[c]{@{}c@{}}Instruction\\ Abbrivation\end{tabular}} & \multicolumn{2}{c}{Description} \\ \hline \hline
\multicolumn{1}{c|}{AR-MI} & \multicolumn{2}{l}{Add from Reference Summary (AR) | Mentioned in Instruction (MI)} \\
\multicolumn{1}{c|}{AR-NMI} & \multicolumn{2}{l}{Add from Reference Summary (AR) | Not Mentioned in Instruction (MI)} \\
\multicolumn{1}{c|}{AA-MI} & \multicolumn{2}{l}{Add from Article | Mentioned in Instruction (MI)} \\
\multicolumn{1}{c|}{OR-MI} & \multicolumn{2}{l}{Omit from Reference Summary (OR) | Mentioned in Instruction (MI)} \\
\multicolumn{1}{c|}{OR-NMI} & \multicolumn{2}{l}{Omit from Reference Summary (OR) | Not Mentioned in Instruction (MI)} \\
\multicolumn{1}{c|}{OA-MI} & \multicolumn{2}{l}{Omit from Reference Summary (OR) | Not Mentioned in Instruction (MI)} \\ \hline 
\end{tabular}
}
\label{tab:edit_types}
\end{wraptable}

To quantify the quality of the hallucination instructions generated by our proposed pipeline, we use two domain expert medical students to annotate the generated instructions and hallucinated summaries with (1) Edit type (Listed in Table \ref{tab:edit_types}), (2) Edit Instruction hallucination label (0=Hallucination instruction, 1=Not and hallucination instruction), and (3) Comment to justify their reasoning for the annotated hallucination label. Detailed human evaluation guidelines are described in \textbf{Appendix \ref{appendixC}}.
Using the annotations from the human evaluators, we focus on the following aspects to quantify the correctness and diversity of the imitation edit instructions and generated hallucinated summaries:
\paragraph{Types of edits in the hallucinated summary} From our evaluation we observed that there were majorly 7 types of edit made from our pipeline as shown in Table \ref{tab:edit_types}. These edits are categorized mainly based on (1) the operation used for the edit (ADD/OMIT), (2) whether the edit was instructed in the instruction generated by our pipeline , and (3) whether the edit was made using the contents from the reference summary or the article (clinical note).  
Figure \ref{fig:HE} (First) illustrates the percentage of edits made in the human evaluation samples for each edit type (Table \ref{tab:edit_types}).  We observe that the majority of hallucination edits made by our pipeline were mentioned in the instructions and were either to add content from the article (AA-MI 34\%) or omit content from the reference summary (OR-MI 21\%). A small proportion of the edits were not mentioned in the generated instruction (AR-NMI 6\% \& OR-NMI 8\%), which we believe was because these edits instructions improve the factuality in the edited summary because of the addition of critical content from the reference summary as well as excluding useless content from the clinical note article. Since we constrained the pipeline to make the edited summary of approximately the same length as the reference summary, so we observed that there was some useful content that was still added from the reference summary (AR-MI 15\%) and some useless content that was omitted from the article (OA-MI 16\%). In conclusion, the edited summary consists of factually important and unimportant content, a high-quality synthetic edit data.  
\paragraph{Imitation edit instructions for hallucination} Focusing on the edits mentioned in the generated instructions, we observed that the majority of edits that led to hallucinations in the edited summary as annotated by the annotators by omitting useful content from the reference summary (OR-MI 49\%), as shown in Figure \ref{fig:HE} (Second). The other hallucinations were made by either adding/omitting useless/useful content from the article with approximately equal proportions (AA-MI 14\% \& OA-MI 36\%). The annotations also showed that none of the hallucination edits in our pipeline were made by adding content from the reference summary (AR-MI 0\%).  
\paragraph{Contribution of ADD/OMIT instructions for hallucination}
\begin{wrapfigure}{r}{0.38\textwidth}
  \begin{center}
    \includegraphics[width=0.37\textwidth]{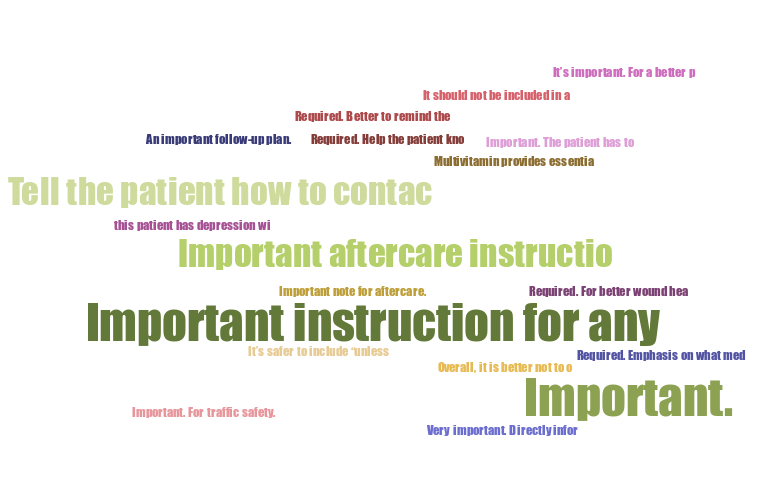}
  \end{center}
  \caption{Word cloud of the comments from the annotators illustrating the reason why they consider an edit instruction to generate hallucination}
  \label{fig:wordcloud}
\end{wrapfigure}
We also investigated the use of ADD $vs$ OMIT operations for generating hallucination edits, as shown in Figure \ref{fig:HE} (Third). According to the annotators, both OMIT operations contribute more wrt ADD operation towards generating hallucination edits (ADD 0\% \& OMIT 39\%), and a majority of ADD operations are responsible for generating edits that do not result in hallucinations (ADD 42\% \& OMIT 18\%). 

From these figures, we conclude that according to the annotators, the majority of edit instructions generated by our pipeline are considered to produce hallucinations in the edited summary, and not only is our pipeline able to generate a variety of edits (listed in Table \ref{tab:edit_types}), but also the majority of edit instructions generated by our pipeline result in hallucinations in the edited summary. Additionally, from Figure \ref{fig:wordcloud}, we illustrate the main comments on why the domain expert annotators believed an edit instruction is responsible for generating hallucination. Their response reasons further validate the quality of our edits for generating hallucinations as the majority of the time they hallucinate content that is usually an \textit{"Important/Required instruction/artifact"} for the discharge summaries to be factually consistent. 


\subsection{External Evaluation}
\setlength\intextsep{0pt}
\begin{wraptable}{R}{7.0cm}
\centering
\small
\caption{External Evaluation Results}
\resizebox{0.5\textwidth}{!}{%
\begin{tabular}{c|ccc|ccc}
\hline
& \multicolumn{3}{c|}{ROUGE} & \multicolumn{3}{c}{Factuality} \\ \cline{2-7} 
& R1 & R2 & RL & G-Eval & UMLS-F1 & Human Rank$\downarrow$ \\ \hline\hline
GPT2-SFT & 30.80 & 13.84 & 23.05 & 5.085 & 20.76 & 3.078 \\
GPT2-DPO & 32.64 & \textbf{15.81} & \textbf{24.51} & 5.492 & \textbf{22.91} & 2.578 \\ 
LLaMA2-SFT & 31.92 & 13.25 & 22.57 & 6.351 & 20.16 & 2.242\\
LLaMA2-DPO & \textbf{33.28} & 14.56 & 24.41 & \textbf{6.648} & 22.05 & \textbf{2.085}\\ 

\hline

\end{tabular}%
}
\label{tab:result}
\end{wraptable}
To evaluate the effectiveness of our proposed edits for improving factuality in the model-generated outputs, we compare the summarization performance of a model trained using a simple fine-tuning (STF) approach optimizing on $\ell_ce$ $vs$ the preference-based DPO training approach that optimizes $\ell_{dpo}$ and uses our proposed synthetic imitation edit pipeline for generating preference data. In simple SFT, the input to the model is the clinical note and the objective is to increase the likelihood of the reference summary, whereas in the case of DPO, the input is the same but the objective is to increase the likelihood of the reference summary while decreasing the likelihood of the hallucinated summary. We experiment with GPT-2 and LLaMA-2 and evaluate the quality of the trained models for summarizatio using ROUGE score (R1,R2,RL), and for factuality using  G-Eval \& UMLS-F1.  
G-Eval evaluates factual alignment using GPT-4 chain-of-thought to assess the factuality when prompted to generate a factuality score (\textbf{Appendix \ref{appendixC}}), whereas UMLS-F1 calculates the F1-score between the UMLS CUI (Unified Medical Language System Controlled Unclassified Information) codes present in the reference summary and the generated summary. For both G-Eval and UMLS-F1, the higher the score/F1, the higher the factuality in the generated output.

From Table \ref{tab:result}, we observe that not only does the use of preference-based DPO training using our proposed synthetic data result in significant improvement in the ROUGE scores, but also there is consistent improvement over the factuality metrics indicating the effectiveness of our proposed pipeline. We also asked the annotators to compare the factuality of the summaries generated by both the models using SFT \& DPO training, resulting in the highest win-rate by the summaries generated using LLaMA2-DPO and the least by GPT2-SFT (both the DPO are preferred over the SFT counterparts).




\section{Conclusion}
In this paper, our study demonstrates the efficacy of utilizing synthetic edit feedback to enhance factual alignment in LLMs for clinical note summarization. We introduce a novel pipeline for generating synthetic preference-based data. Additionally, human annotators consistently favored the factuality of summaries generated through our approach.

\section{Limitations}
In this paper, we focus only on the task factuality alignment in the clinical summarization, and adapting the proposed method to other domains is yet to be explored. Since we only had access to medical students as annotators, we relied on them for one of our factuality metrics (Human Rank) for quantifying the preference results. Although they are qualified to read and annotate clinal notes and their corresponding discharge summaries, using more qualified domain experts as annotators would further increase the statistical significance of our results, which we leave to future work. We further leave it to future work to address concerns about fairness, generalizability to other domains/languages, and potential biases inherent in LLMs.

\bibliography{main,anthology}
\bibliographystyle{acl_natbib}

\appendix
\section{Hallucination Edit Prompt} \label{appendixA}

\begin{longtable}{ p{12cm} }
\caption{Edit Prompt}\\
    \hline
    \textbf{»»»» Instruction »»»»} \\
    You are a clinical writing assistant who is in edit mode. You are tasked with generating hallucinated summary based on provided a clinical note article and a reference summary for the article. The goal is to edit the reference summary to generate a hallucinated summary that sounds plausible but includes edits introduced through an edit operation which can be one of the following: 
    \paragraph{Add Operation:} Intentionally add medico-legally essential words from the article not required for accurate diagnosis and treatment documentation.
    \paragraph{Omit Operation:} Intentionally omit medico-legally essential words in the reference summary required for accurate diagnosis and treatment documentation. \\ \\
    For these operations focus on words that, if missing or incorrect in the hallucinated summary, could lead to wrong diagnoses and treatments in the future. Maintain coherence while excluding essential terms. The hallucinated summary should be concise and contain no more than FIVE EXTRA WORDS compared to the reference summary and should have an equal number of Add/Omit operations. \\ \\
    Steps for generating the hallucinated summary:
    \paragraph{Step 1:} List the proposed edit operations to introduce hallucination on the reference summary.
    \paragraph{Step 2:} Use the proposed edit operations to edit the reference summary. \\ \\
    \textbf{»»»» Output Format »»»»} \\
    The output format is: \\
           Numbererd List hallucination edits made:\\
           \{Edit 1\}, \{Edit 2\}, \{Edit 3\} ...\\
           Hallucinated Summary: \\  \\
    \textbf{»»»» Follow the above Instructions, Hallucination Method and Output Format »»»»}\\
	Now, let's start.\\
	Generate the hallucinated summary:\\
	Article - \{src\}\\
	Reference Summary - \{ref\} \\
    \hline

\label{fig:edit-prompt}
\end{longtable}

\section{Human Evaluation Annotation Guidelines}
\label{appendixB}

\begin{table}[ht]
\centering
\caption{Human Evaluation Guideline}

\resizebox{0.7\textwidth}{!}{%
\begin{tabular}{cll}
\hline
\multicolumn{3}{c}{Hallucination Instruction Identification Guideline for Add/Omit Operations} \\ \hline
\multicolumn{1}{c|}{Op.} & \multicolumn{1}{c|}{Label} & \multicolumn{1}{c}{Description} \\ \hline \hline
\multicolumn{1}{c|}{ADD} & \multicolumn{1}{l|}{0} & \begin{tabular}[c]{@{}l@{}}Including medico-legally phrases from the Article/Reference Summary \\ that are required for accurate diagnosis and treatment documentation.\end{tabular} \\ \cline{3-3}
\multicolumn{1}{c|}{ADD} & \multicolumn{1}{l|}{1} & \begin{tabular}[c]{@{}l@{}}Including medico-legally phrases from the Article/Reference Summary \\ that are not required for accurate diagnosis and treatment documentation.\end{tabular} \\ \cline{3-3}
\multicolumn{1}{c|}{OMIT} & \multicolumn{1}{l|}{0} & \begin{tabular}[c]{@{}l@{}}Not Including medico-legally phrases from the Article/Reference Summary\\ that are not required for accurate diagnosis and treatment documentation.\end{tabular} \\ \cline{3-3}
\multicolumn{1}{c|}{OMIT} & \multicolumn{1}{l|}{1} & \begin{tabular}[c]{@{}l@{}}Not Including medico-legally phrases from the Article/Reference Summary \\ that are required for accurate diagnosis andtreatment documentation.\end{tabular} \\ \hline
\end{tabular}%
}
\label{tab:human-eval-guideline}
\end{table}

For the human evaluation, we provided the annotators with a set of clinical note articles (article), reference summaries with the corresponding discharge instructions with reference to the article, and a list of edit instructions (edit instructions). The edit instructions consisted of two operations (Add \& Omit operations) using which a new summary called edited summary can be generated. 

The two operations are described below:
\begin{enumerate}
    \item \textbf{Add Operation}: Intentionally including medico-legally phrases in the edited summary from the article or reference summary that are not required for accurate diagnosis and treatment documentation.
    \item \textbf{Omit Operation}: Intentionally not including medico-legally phrases in the edited summary from the article or reference summary that are required for accurate diagnosis and treatment documentation.
\end{enumerate}

Both the above operations in the edit instruction can be used to generate hallucinations in the edited summary, where hallucinations are the phrases that are either (1) not present in the edited summary that is crucial for accurate diagnosis and treatment documentation, or (2) present in the edited summary that are not crucial for accurate\ diagnosis and treatment documentation. Edit instruction that leads to hallucination is called a hallucination instruction.

The condition for an edit instruction with either Add or Omit operation is a hallucination instruction is listed in Table \ref{tab:human-eval-guideline}. In the table above, the hallucination label is used to label if an instruction leads to hallucination in the edited summary or not (0=Hallucination instruction, 1=Not and hallucination instruction).

Given the article, reference summary, and edit instructions generated by our pipeline, we asked the annotators to annotate each instruction with its:
\begin{enumerate}
    \item Hallucination Label: 0 if the instruction is a hallucination instruction or 1 if not.
    \item Comment: Justification for the hallucination label.
\end{enumerate}
Further, we also asked them to annotate the edit type listed int Tabel \ref{tab:edit_types} (Section \ref{section4_1}) for each edit in the generated hallucinated summary.

\section{G-Eval Factuality Metric Prompt} \label{appendixC}
\begin{longtable}{ p{12cm} }
\caption{G-Eval Factuality Prompt}\\
    \hline
    You will be given one discharge summary written for a Clinical Note. \\
    Your task is to rate the summary on one metric.
    Please make sure you read and understand these instructions carefully. Please keep this document open while reviewing, and refer to it as needed. \\
    \textbf{Evaluation Criteria:}\\
    Factual Consistency (1-10): Is the summary has missing or incorrect facts that are not supported by the source text and could lead to wrong diagnoses and treatments? \\
    \textbf{Evaluation Steps:}\\
    \begin{enumerate}
        \item Read the clinical note carefully and identify the main topic and key points.
        \item Read the discharge summary and compare it to the clinical notee. Check if the summary covers the main topic and key points of the  clinical note, and Is the summary has missing or incorrect facts that are not supported by the source text and could lead to wrong diagnoses and treatments?
        \item Assign a score for Factual Consistency on a scale of 1 to 10, where 1 is the lowest and 10 is the highest based on the Evaluation Criteria.
    \end{enumerate}
    \textbf{Clinical Note Text:}\\
    \{Document\}\\
    \textbf{Reference Discharge Summary:}\\
    \{Reference Summary \}\\
    \textbf{System Output Discharge Summary:}\\
    \{System Output Summary\}\\ \\
    Return the scores as dictionary objects, adhering to the following structure:\\
    \{"Factual Consistency": ...\} \\ 
    Please provide your response solely in the dictionary format without including any additional text. \\
    \hline

\label{fig:geval-prompt}
\end{longtable}

\section{Human Evaluation Examples}
\label{appendixD}
\begin{figure}[ht]
  \centering
    \includegraphics[width=0.7\textwidth]{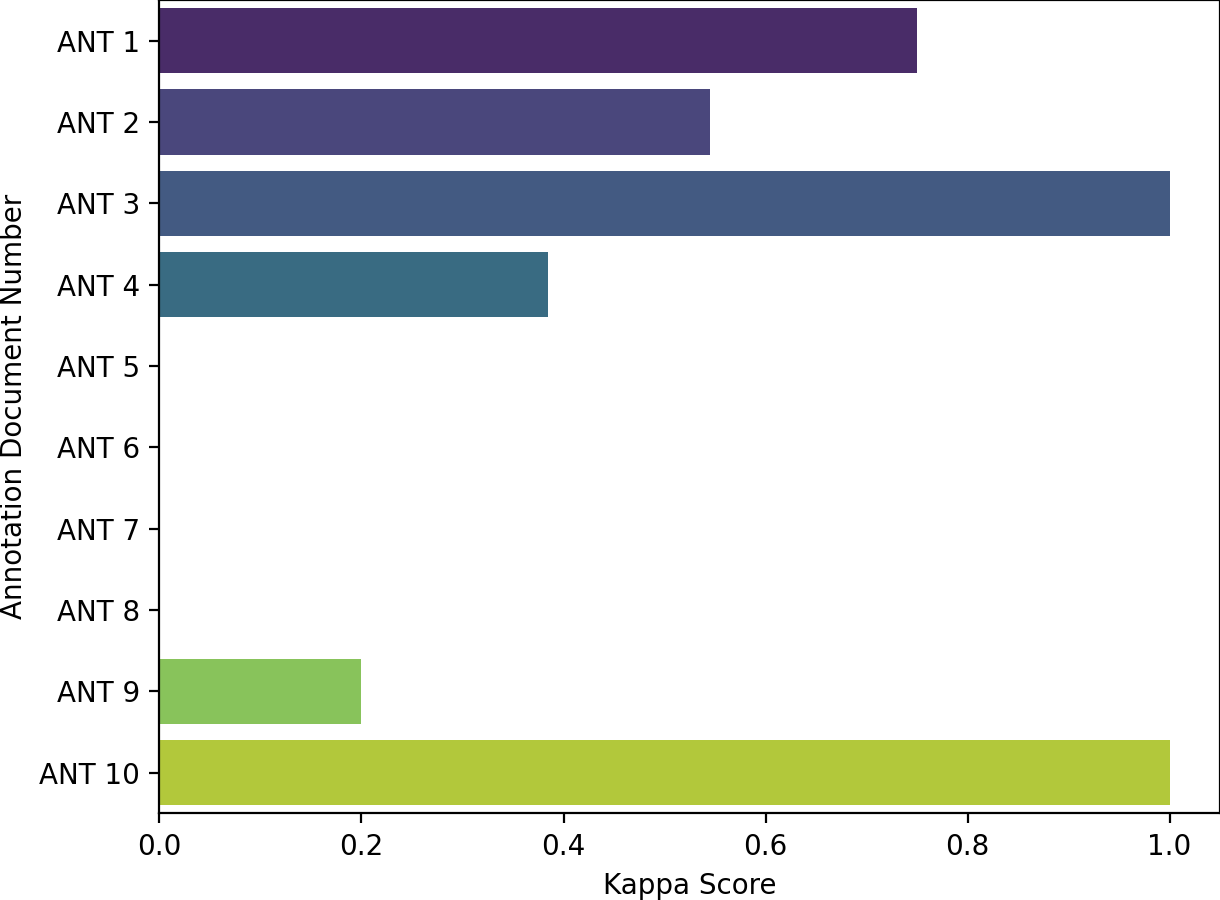}
  \caption{Kappa score for all annotation documents (ANN \#) used in the human evaluation}
  \label{fig:annotate}
\end{figure}

For our human evaluation, we also calculated the Kappa score for inter-annotator agreement for hallucination labels in each document annotated by our annotators. We observed a mean Kappa score of 0.38, and the Kappa score plot for each of the documents is shown in Figure \ref{fig:annotate}. We observe that for the majority of the documents, the annotators were in high agreement with each other for the hallucination label, but in some documents in which the ADD/OMIT operation was being done using the contents of the article instead of the reference, there was a high degree of disagreement for deciding the hallucination label. We provide two examples for our human annotation in Table \ref{tab:example1} \& Table \ref{tab:example2}. 

\begin{longtable}{ p{12cm} }
\caption{Human Annotation No. 1} \\ 
    \hline
    Clinical Note  \\ \hline
    \textbf{Brief Hospital Course:} Pt was given 3 units of PRBCs and 4L NS for resusitation in the ED and admitted to the MICU where he was NPO, having serial HCT q4hr, a 4th unit of PRBCs, [**Hospital1 **] IV PPI, and close monitoring with 2 large bore IVs. Pt got an EGD which showed 2 ulcers, 1x1am and 1x2cm near the anastomosis from his prior gastric bypass. The larger ulcer had an adherent clot underneath which there was a visible vessel and more clot. BICAP was applied to cauterize. There was no bleeding noted at the conclusion of the EGD. KUB and CXR post-procedure revealed no free air in the abdomen. Pt's HCT went from 21.8 on admission to 26 after his transfusions. Pt remained NPO with maintenance IVF, electrolyte replacement, and serial HCT until HD\# 3. Pt had no further bleeding and was hemodynamically stable and was transferred to a monitored floor bed on HD\#3. The pt began taking sips and then clears by mouth on HD\#3. He was able to take in a Stage III diet and oral PPI medication on HD\#4 and his HCT was stable at 27.6 and 27.4. The pt was sent home on a PPI and instructed to f/u with his surgeon within a week or to call sooner if her experienced any further bleeding.
    \\ \hline
    Reference Summary   \\ \hline
    \textbf{Discharge Instructions:} Please call you doctor if you experience any of the following: - fever >101 - blood with bowel movements or blood in vomit - dizziness or lightheadedness - persistent nausea and vomiting - inability to eat or drink - severe abdominal pain - any other questions or concerns Please take your medications as prescribed. Please also take a chewable multivitamin, like Flintstones, daily. Please follow up with Dr. [**Last Name (STitle) **]. Please remain on Stage III diet until you follow up with your doctor. Do not self-advance your diet. Also, do not chew gum or drink out of a straw. \\ \hline
    Edit Instructions   \\ \hline
    Numbered List hallucination edits made: \\
    \begin{enumerate}
        \item Add Operation: "Please call your doctor if you experience any of the following:"
        \item Omit Operation: "- any other questions or concerns"
        \item Add Operation: "Please take your medications as prescribed."
        \item Omit Operation: "Please also take a chewable multivitamin, like Flintstones, daily."
        \item Add Operation: "Please follow up with Dr. [**Last Name (STitle) **]."
        \item Omit Operation: "Please remain on Stage III diet until you follow up with your doctor."
        \item Add Operation: "Do not self-advance your diet."
        \item Omit Operation: "Also, do not chew gum or drink out of a straw."'
    \end{enumerate}
    \\ \hline
    Hallucinated Summary   \\ \hline
    \textbf{Discharge Instructions:} Please call your doctor if you experience any of the following: - fever >101 - blood with bowel movements or blood in vomit - dizziness or lightheadedness - persistent nausea and vomiting - inability to eat or drink - severe abdominal pain. Please take your medications as prescribed. Please follow up with Dr. [**Last Name (STitle) **]. Do not self-advance your diet.
    \\ \hline
    Human Annotation \#1 \\ \hline
    Instruction 1: \\
    Hallucination Label: 0 \\
    Comment: Important instruction for any potential emergencies or progression. Required. \\
    Instruction 2: \\
    Hallucination Label: 1 \\
    Comment: Important instruction for any potential emergencies or progression. Required.  \\
    Instruction 3: \\
    Hallucination Label: 0 \\
    Comment: Required.  \\
    Instruction 4: \\
    Hallucination Label: 1 \\
    Comment: Multivitamin provides essential components required to produce new blood cells. Required.  \\
    Instruction 5:\\
    Hallucination Label: 0 \\
    Comment: Tell the patient how to contact. Required. \\
    Instruction 6: \\ 
    Hallucination Label: 1 \\
    Comment: Important note for aftercare.  \\
    Instruction 7: \\
    Hallucination Label: 0 \\
    Comment: Emphasis on instruction 6.  \\
    Instruction 8: \\ 
    Hallucination Label: 1 \\
    Comment: Chewing gum may stimulate gastric acid to secrete. Drinking through a straw may increase the pressure in the cavity of upper GI tract and consequently trigger rebleeding, which may be controversial. Overall, it is better not to omit these two suggestions.  \\
    \\ \hline
    Human Annotation \#2 \\ \hline
    Instruction 1: \\
    Hallucination Label:1 \\
    Comment: more detail needed \\
    Instruction 2: \\
    Hallucination Label:1 \\
    Comment:very simply \\
    Instruction 3: \\
    Hallucination Label:1 \\
    Comment: Useful doctor's advice \\
    Instruction 4: \\
    Hallucination Label:0 \\
    Comment: Useful doctor's advice \\
    Instruction 5: \\
    Hallucination Label:1 \\
    Comment: It adds more detail \\
    Instruction 6: \\
    Hallucination Label:0 \\
    Comment: Useful doctor's advice \\
    Instruction 7: \\
    Hallucination Label:1 \\
    Comment: Useful doctor's advice \\
    Instruction 8: \\
    Hallucination Label:0 \\
    Comment: Useful doctor's advice \\
    \\ \hline

\label{tab:example1}
\end{longtable}

\begin{longtable}{ p{12cm} }
\caption{Human Annotation No. 2} \\ 
    \hline
    Clinical Note  \\ \hline
    \textbf{Brief Hospital Course:} Pt was admitted after catherization after IV hydration. On [**2155-2-13**] she went to the operating room where she underwent an aortic valve replacement with size 23-mm St. [**Male First Name (un) 923**] Epic tissue valve. See operative note for full details. Overall the patient tolerated the procedure well and post-operatively was transferred to the CVICU in stable condition for recovery and invasive monitoring. POD 1 found the patient extubated, alert and oriented and breathing comfortably. The patient was neurologically intact and hemodynamically stable on no inotropic or vasopressor support. Low dose beta blocker was initiated but then stopped due to hypotension with a SBP in the 80's. The patient was gently diuresed toward the preoperative weight. The patient was transferred to the telemetry floor for further recovery. Chest tubes were split due to high drainage and mediatinal chest tubes were removed with subsequent removal of left pleural chest tube once drainage had decreased. She was anticoagulated with Coumadin for chronic atrial fibrillation. The patient was evaluated by the physical therapy service for assistance with strength and mobility. By the time of discharge on POD \#7 the patient was ambulating with assistance, the wound was healing well and pain was controlled with Tylenol only due to somnolence with Ultram. The patient was discharged to [**Hospital **] in [**Location (un) 246**] in good condition with appropriate follow up instructions.First INR check day after discharge. Target INR 2.0-2.5 .
    \\ \hline
    Reference Summary   \\ \hline
    \textbf{Discharge Instructions:} Please shower daily including washing incisions gently with mild soap, no baths or swimming until cleared by surgeon. Look at your incisions daily for redness or drainage Please NO lotions, cream, powder, or ointments to incisions Each morning you should weigh yourself and then in the evening take your temperature, these should be written down on the chart No driving for approximately one month and while taking narcotics, will be discussed at follow up appointment with surgeon when you will be able to drive No lifting more than 10 pounds for 10 weeks Please call with any questions or concerns [**Telephone/Fax (1) 170**] Females: Please wear bra to reduce pulling on incision, avoid rubbing on lower edge **Please call cardiac surgery office with any questions or concerns [**Telephone/Fax (1) 170**]. Answering service will contact on call person during off hours** \\ \hline
    Edit Instructions   \\ \hline
    Numbered List hallucination edits made: \\
    \begin{enumerate}
        \item Add: Pt was admitted after catherization after IV hydration.
        \item Omit: Please shower daily including washing incisions gently with mild soap, no baths or swimming until cleared by surgeon.
        \item Add: Look at your incisions daily for redness or drainage.
        \item Omit: Please NO lotions, cream, powder, or ointments to incisions.
        \item Add: Each morning you should weigh yourself and then in the evening take your temperature, these should be written down on the chart.
        \item Omit: No driving for approximately one month and while taking narcotics, will be discussed at follow up appointment with surgeon when you will be able to drive.
        \item Add: The patient was evaluated by the physical therapy service for assistance with strength and mobility.
        \item Omit: No lifting more than 10 pounds for 10 weeks.
        \item Add: The patient was discharged to [**Hospital **] in [**Location (un) 246**] in good condition with appropriate follow up instructions.
        \item Omit: Please call with any questions or concerns [**Telephone/Fax (1) 170**].
        \item Add: Females: Please wear bra to reduce pulling on incision, avoid rubbing on lower edge.
        \item Omit: **Please call cardiac surgery office with any questions or concerns [**Telephone/Fax (1) 170**]. Answering service will contact on call person during off hours**.

    \end{enumerate}
    \\ \hline
    Hallucinated Summary   \\ \hline
    \textbf{Discharge Instructions:} Pt was admitted after catherization after IV hydration. Look at your incisions daily for redness or drainage. Each morning you should weigh yourself and then in the evening take your temperature, these should be written down on the chart. The patient was evaluated by the physical therapy service for assistance with strength and mobility. The patient was discharged to [**Hospital **] in [**Location (un) 246**] in good condition with appropriate follow up instructions. Females: Please wear bra to reduce pulling on incision, avoid rubbing on lower edge.
    \\ \hline
    Human Annotation \#1 \\ \hline
    Instruction 1: \\
    Hallucination Label: 1 \\
    Comment: It should not be included in a discharge instruction. But it’s necessary as a part of the clinical note.  \\
    Instruction 2: \\
    Hallucination Label: 1 \\
    Comment: Important aftercare instruction. Helps prevent infection.  \\
    Instruction 3: \\
    Hallucination Label: 0 \\
    Comment: Important aftercare instruction. Redness or drainage may indicate infection.  \\
    Instruction 4: \\
    Hallucination Label: 1 \\
    Comment: Important aftercare instruction, decreasing the risk of infection or non-healing wound.  \\
    Instruction 5: \\
    Hallucination Label: 0 \\
    Comment: Necessary and required to keep a trace of the recovery course. Weight may indicate volume status (hypovolemia, hypervolemia, or euvolemia) which helps judge heart function. Elevated temperature is associated with infection.  \\
    Instruction 6: \\
    Hallucination Label: 1 \\
    Comment: Important. For traffic safety. Because narcotic medications may sedate the patient, disturbing judgment and agility.  \\
    Instruction 7: \\
    Hallucination Label: 1 \\
    Comment: Unrelated. Similar to Instruction 1.  \\
    Instruction 8: \\
    Hallucination Label: 1 \\
    Comment: Required. For better wound healing and overall recovery.  \\
    Instruction 9: \\
    Hallucination Label: 1 \\
    Comment: Unrelated. Similar to Instruction 1.  \\
    Instruction 10: \\
    Hallucination Label: 1 \\
    Comment: Tell the patient how to contact. Required.  \\
    Instruction 11: \\
    Hallucination Label: 0  \\
    Comment: Required aftercare. For better incision healing. \\ 
    Instruction 12: \\
    Hallucination Label: 1 \\
    Comment: Tell the patient how to contact. Required. \\ \hline
    Human Annotation \#2 \\  \hline
    Instruction 1: \\
Hallucination Label:0 \\
Comment: makes summary more accurate \\
Instruction 2: \\
Hallucination Label:0 \\
Comment:should be more accurate \\
Instruction 3: \\
Hallucination Label:0 \\
Comment: Increase doctor's orders and reduce postoperative complications \\
Instruction 4: \\
Hallucination Label:1 \\
Comment:should be more details \\
Instruction 5: \\
Hallucination Label:0 \\
Comment:makes summary more accurate \\
Instruction 6: \\
Hallucination Label:1 \\
Comment:should be more accurate \\
Instruction 7: \\
Hallucination Label:0 \\
Comment:adds more details \\
Instruction 8: \\
Hallucination Label:1 \\
Comment: Medical advice should be increased \\
Instruction 9: \\
Hallucination Label:0 \\
Comment:Adds more details \\
Instruction 10: \\
Hallucination Label:1 \\
Comment: patients need to know how to contact the hospital \\
Instruction 11: \\
Hallucination Label:0 \\
Comment:adds more details \\
Instruction 12: \\
Hallucination Label:1 \\
Comment: patients need to know how to contact the hospital \\ \hline
\label{tab:example2}
\end{longtable}


\end{document}